\documentclass[conference]{IEEEtran}
\IEEEoverridecommandlockouts
\usepackage{cite}
\usepackage{amsmath,amssymb,amsfonts}
\usepackage{algorithmic}
\usepackage{graphicx}
\usepackage{textcomp}
\usepackage{xcolor}
\usepackage{multirow}
\usepackage{makecell} 
\usepackage{float}
\usepackage{amsfonts}
\usepackage{times}  
\usepackage[bottom]{footmisc}  
\renewcommand{\footnotesize}{\fontsize{9pt}{9pt}\selectfont}  
\usepackage{booktabs}  
\usepackage{placeins}   
\usepackage{stfloats}  
\usepackage{lipsum} 
\def\BibTeX{{\rm B\kern-.05em{\sc i\kern-.025em b}\kern-.08em
    T\kern-.1667em\lower.7ex\hbox{E}\kern-.125emX}}
\begin{document}

\title{Shifting Spotlight for Co-supervision: A Simple yet Efficient Single-branch Network to
See Through Camouflage
\thanks{Corresponding author\IEEEauthorrefmark{1}. 
This work was supported by Research Fund for Advanced Ocean Institute of Southeast University, Nantong (GP202411), Guangdong Basic and Applied Basic Research Foundation (2022A1515011435), ZhiShan Scholar Program of Southeast University and the Fundamental Research Funds for the Central Universities, the Natural Science Basic Research Program of Shaanxi (Program No.2024JC-YBMS-513), and Key Research and Development Program of Zhejiang Province under Grants 2024C01025.}
}

\author{
    \IEEEauthorblockN{
        Yang Hu$^{1}$, 
        Jinxia Zhang\IEEEauthorrefmark{1}$^{1,3}$, 
        Kaihua Zhang$^{2}$, 
        Yin Yuan$^{1}$, 
        Jiale Huang$^{1}$,
        Zechao Zhan$^{1}$,
        Xin Wang$^{4}$ 
    }
    \IEEEauthorblockA{
        $^{1}$\textit{Key Laboratory of Measurement and Control of CSE, School of Automation, Southeast University, China}\\
       $^{2}$\textit{School of Computer Science, Nanjing University of Information Science and Technology, China}\\
       $^{3}$\textit{Advanced Ocean Institute of Southeast University, China} $^{4}$\textit{Alibaba Group, China}\\
    }
}

\maketitle

\begin{abstract} 
Camouflaged object detection (COD) remains a challenging task in computer vision. Existing methods often resort to additional branches for edge supervision, incurring substantial computational costs. To address this, we propose the Co-Supervised Spotlight Shifting Network (CS\(^3\)Net), a compact single-branch framework inspired by how shifting light source exposes camouflage. Our spotlight shifting strategy replaces multi-branch designs by generating supervisory signals that highlight boundary cues. Within CS\(^3\)Net, a Projection Aware Attention (PAA) module is devised to strengthen feature extraction, while the Extended Neighbor Connection Decoder (ENCD) enhances final predictions. Extensive experiments on public datasets demonstrate that CS\(^3\)Net not only achieves superior performance, but also reduces Multiply-Accumulate operations (MACs) by 32.13\% compared to state-of-the-art COD methods, striking an optimal balance between efficiency and effectiveness.
\vspace{-1em}
\end{abstract}

\section{Introduction}
\label{sec:intro}
Camouflage is an evolutionary adaptation that allows animals to evade predators’ attention by blending into the surroundings. This natural phenomenon has intrigued scientists and inspired diverse practical applications, such as detecting medical lesions \cite{fan2020pranet,chiao2011hyperspectral,ji2021progressively,ji2022video,fan2020inf,wu2021jcs,liu2021covid,chen2022pedestrian} and identifying industrial defects \cite{bergmann2018defect1,tabernik2020defect2,tsai2021defect3}. Camouflaged Object Detection (COD) is particularly difficult because of the visually deceptive nature of camouflage. It often relies on the use of computationally intensive networks for accurate detection. Therefore, developing an efficient and effective COD method is of significant importance. 
\par The major challenge of precisely segmenting camouflaged targets lies in the blurred boundaries between them and the surroundings. Recent deep learning based methods \cite{edge_aware_SCRN,errnet,mutualGL,edge_reconstruion,edge_aware_mirrornet} effectively tackle this challenge by introducing an additional branch to integrate extra prior information (such as edge, depth, \textit{etc.}). However, the additional branch substantially increases the number of parameters and computational demand, ultimately reducing the overall efficiency.
\begin{figure}[tbp]  
    \centering
    \includegraphics[width=0.98\linewidth]{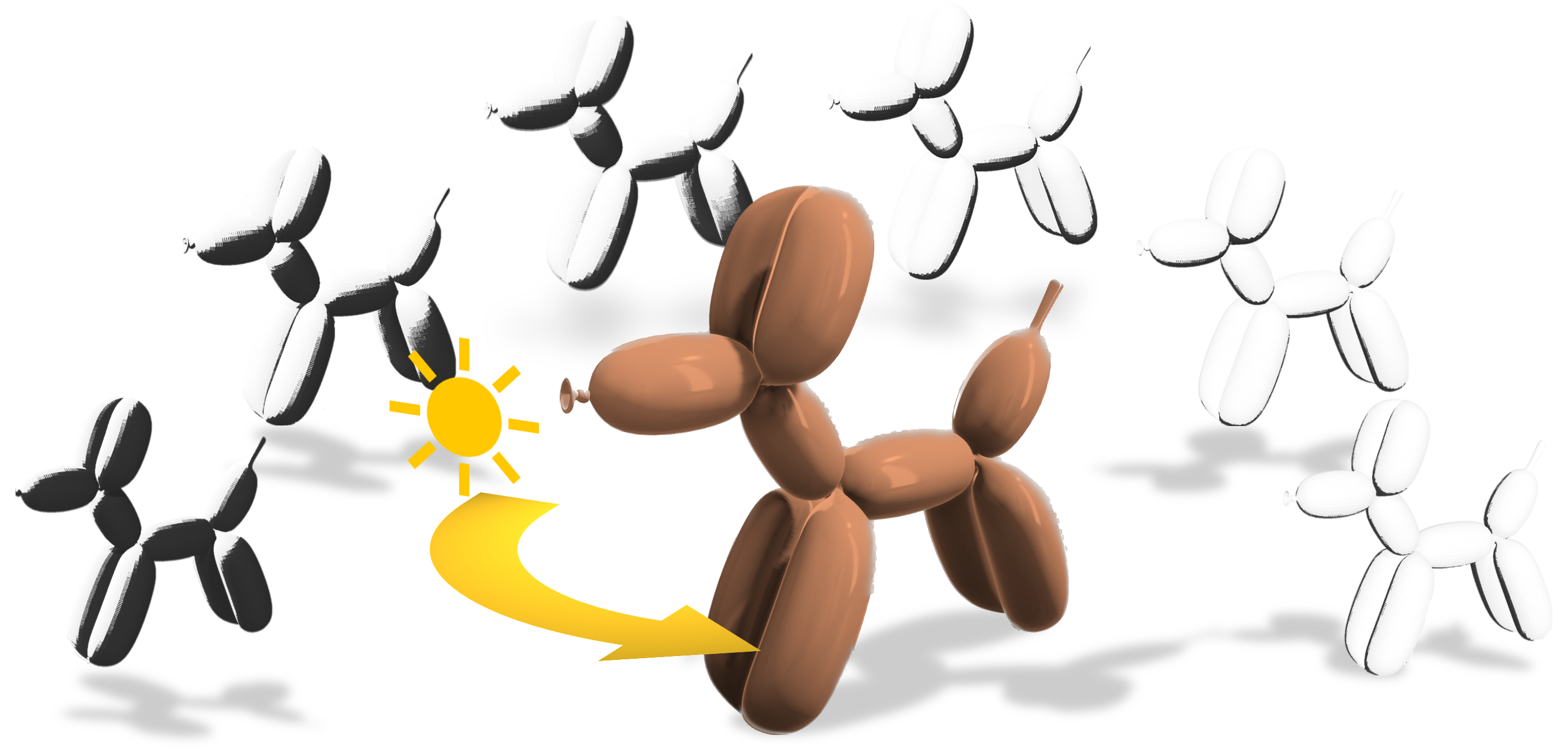}  
    \vspace{-1.1em}
    \caption{An illustration of spotlight shifting strategy,  the shadow projection over the balloon dog changes dramatically as the spotlight shifts from left to right. These changes highlights the object’s contours, making them more distinct and easier to detect.}
    \vspace{-1.5em}
    \label{fig:teaser_spotlight}
\end{figure}
\begin{figure}[!b]
    \vspace{-1.9em}
    \includegraphics[width=0.98\linewidth]{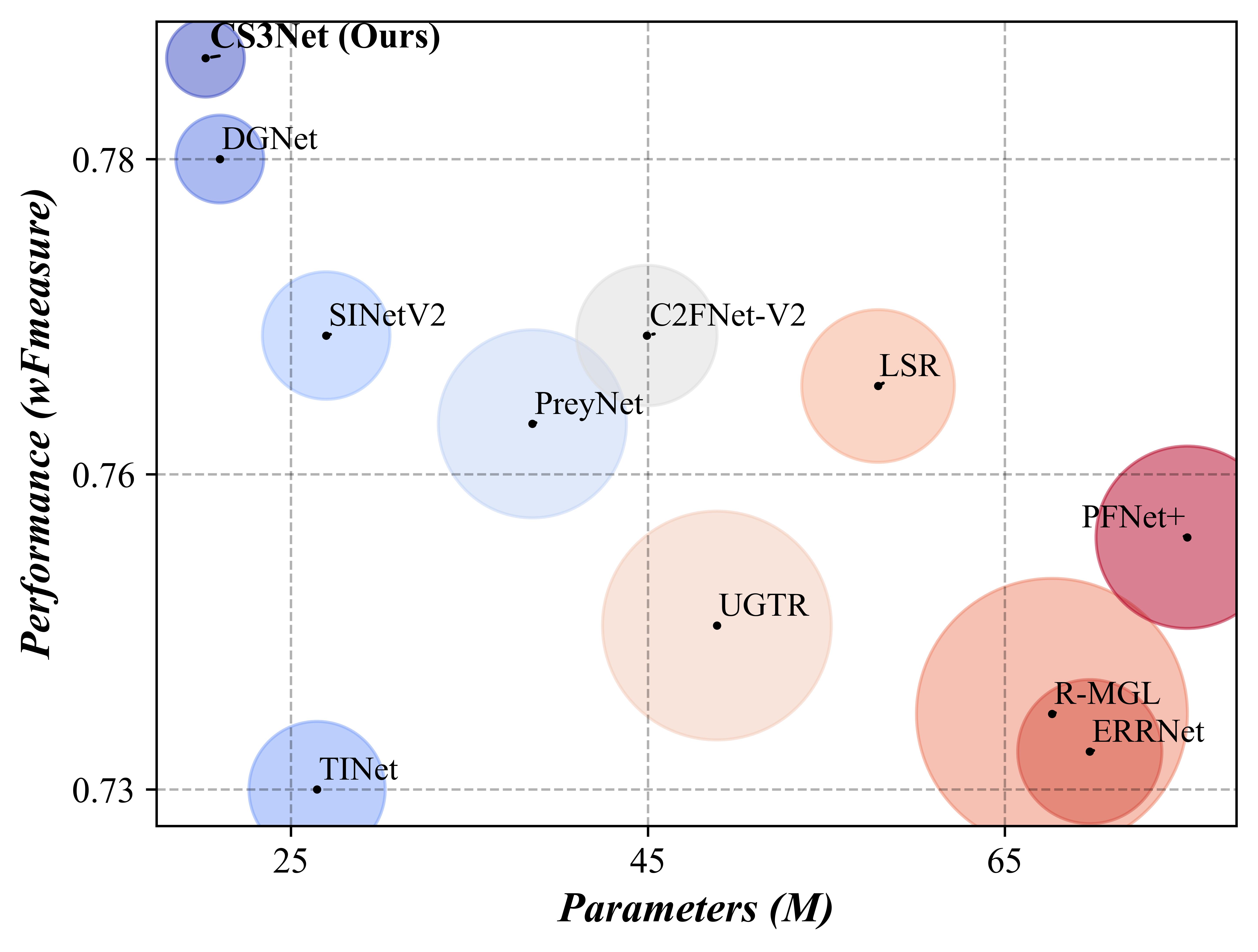}
    \vspace{-1.3em}
    \caption{Comparison of model performance ($F^w_\beta$ on NC4K), parameters and MACs across state-of-the-art COD methods.}
    \label{fig:scatter_chart}
\end{figure}
\par To address the inefficiencies caused by multi-branch structures, we propose a spotlight shifting strategy that enables network co-supervision within a single-branch framework. This strategy is inspired by how varying light condition can make the objects' contours more distinct. As presented in Fig.~\ref{fig:teaser_spotlight}, the shifting light source leads to eye-catching changes on shadow projection, enhancing the visibility of the object. Our proposed spotlight shifting strategy mimics this effect by generating shadow maps with different spotlight positions, which are further used for co-supervision. This strategy allows the model to capture informative cues without extra branches.
\par Our main goal is to develop a COD method that strikes the optimal performance-efficiency trade-off, as evidenced by Fig.~\ref{fig:scatter_chart}. Targeting at which, we present the \textbf{C}o-\textbf{S}upervised \textbf{S}potlight \textbf{S}hifting \textbf{N}etwork (CS\(^3\)Net), a single-branch network that leverages spotlight shifting co-supervision for enhanced feature extraction. It integrates the proposed Projection Aware Attention (PAA) and Extended Neighbor Connection Decoder (ENCD) for feature refinement and precise prediction. Our contributions are summarized as follows:
\begin{itemize}
    \vspace{-0.3em}
    \item We propose spotlight shifting strategy for network co-supervision that enriches the model capability to discern camouflaged objects without introducing extra branches, providing a unique perspective in COD domain.
    \item Based on spotlight shifting strategy, a novel single-branch and efficient model, termed as CS\(^3\)Net is crafted for COD task. It integrates the proposed PAA, and ENCD to precisely enhance feature representations.
    \item CS\(^3\)Net achieves the optimal balance between model efficiency and performance, reducing MACs by 32.13\% compared to the most advanced efficient COD models, while also delivering superior performance.
    \vspace{-0.5em}
\end{itemize}
\begin{figure*}[th]  
    \centering
    \includegraphics[width=\textwidth]{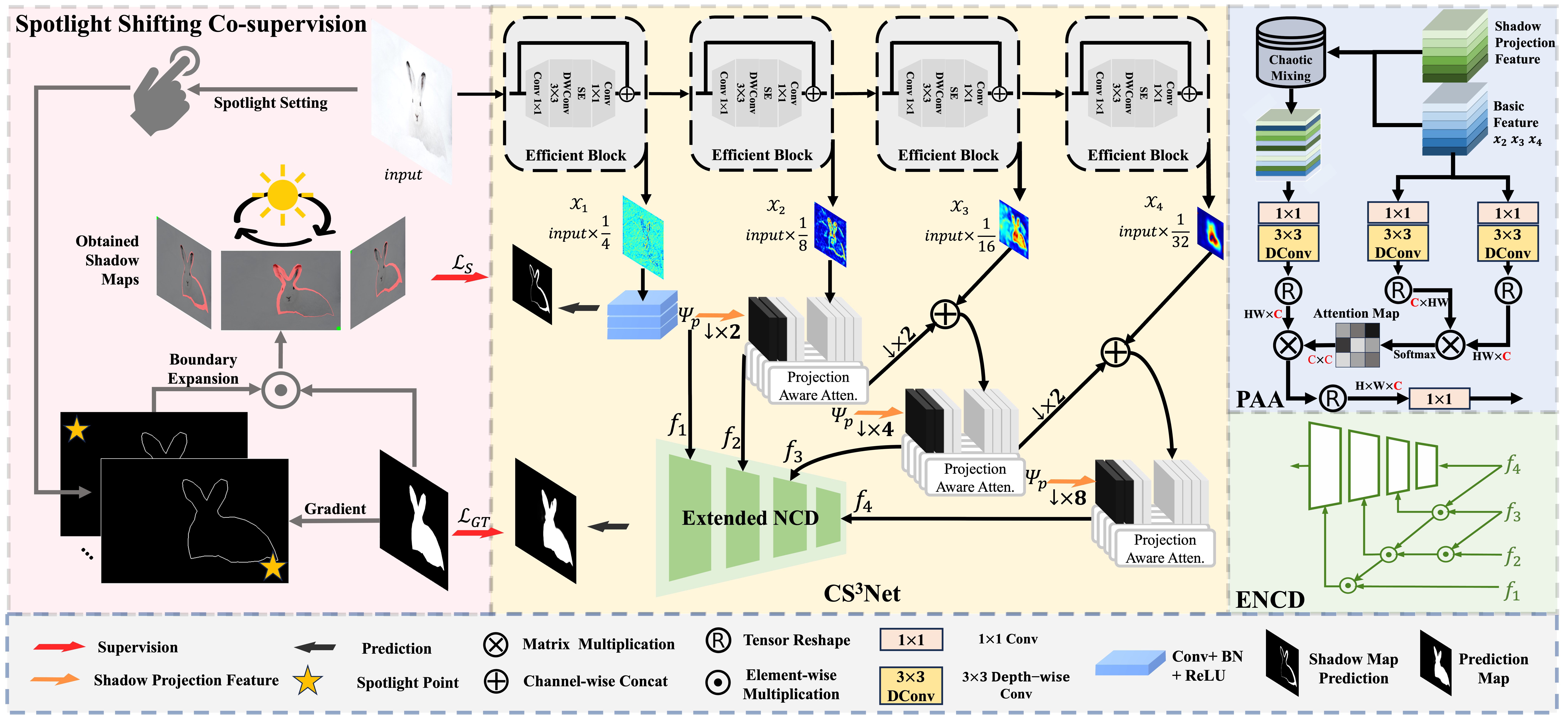}
    \vspace{-2.0em}
    \caption{The architecture of the proposed CS\(^3\)Net. CS\(^3\)Net operates as a single-branch network utilizing spotlight shifting strategy for co-supervision, it consists of two key modules: the Projection Aware Attention (PAA) and the Extended Neighbor Connection Decoder (ENCD) to integrate knowledge gained from co-supervision.}
    \label{fig:network_struct}
    \vspace{-2em}
\end{figure*}
\section{The Proposed Method}
\label{sec:method}
As illustrated in Fig.~\ref{fig:network_struct}, CS\(^3\)Net is a single-branch network that leverages Spotlight Shifting Co-supervision to refine feature extraction. The network first employs EfficientNet \cite{tan2019efficientnet} to capture multi-scale features $\left \{ x_{i}  \right \} _{i=0}^{4}$ , followed by the Projection Aware Attention (PAA) to progressively align attention with shadow projection feature. Finally, the Extended Neighbor Connection Decoder (ENCD) efficiently fuses these refined features to improve prediction accuracy.
\begin{table*}[htp]
\caption{Quantitative comparison on three datasets, with the best scores highlighted in \textbf{bold}. Methods involving efficient design are selectively highlighted for comparison.}
\vspace{-1em}
\centering
\resizebox{0.98\linewidth}{!}
{
\begin{tabular}{rrrrcccccccccccc}
\toprule[1.5pt] 
\multicolumn{1}{r}{\multirow{2}{*}{\textbf{Models}}} &
\multirow{2}{*}{\textbf{Pub./Year}} &
    \multirow{2}{*}{\makecell[c]{\textbf{Params.}\\\textbf{(M) ↓}}} &
    \multirow{2}{*}{\makecell[c]{\textbf{MACs}\\\textbf{(G) ↓}}} &
    \multicolumn{4}{c}{\textbf{NC4K \cite{lv2021lsr_nc4k}}} &
    \multicolumn{4}{c}{\textbf{CAMO \cite{CAMO_Abranch}}} &
    \multicolumn{4}{c}{\textbf{COD10K \cite{COD10K_SINet}}} \\ \cmidrule(lr){5-8}\cmidrule(lr){9-12}\cmidrule(lr){13-16}
\multicolumn{1}{c}{}   & & & & \textbf{$S_\alpha$ ↑}& \textbf{$E_{\phi}$ ↑}& \textbf{$F^w_\beta$ ↑}& \textbf{$MAE$ ↓}& \textbf{$S_\alpha$ ↑}& \textbf{$E_{\phi}$ ↑}& \textbf{$F^w_\beta$ ↑}& \textbf{$MAE$ ↓}& \textbf{$S_\alpha$ ↑}&\textbf{$E_{\phi}$ ↑}& \textbf{$F^w_\beta$ ↑}& \textbf{$MAE$ ↓}\\ \midrule[0.5pt]
SINet \cite{COD10K_SINet}      & CVPR$_{20}$ & 48.95  & 19.42  & 0.808 & 0.871 & 0.723 & 0.058 & 0.745 & 0.804 & 0.644 & 0.092 & 0.776 & 0.864 & 0.631 & 0.043 \\
UGTR \cite{yang2021UGTR}       & ICCV$_{21}$ & 48.87  & 127.12 & 0.839 & 0.874 & 0.747 & 0.052 & 0.785 & 0.823 & 0.686 & 0.086 & 0.818 & 0.853 & 0.667 & 0.035 \\
R-MGL \cite{mutualGL}          & CVPR$_{21}$ & 67.64  & 249.89 & 0.833 & 0.867 & 0.740 & 0.052 & 0.775 & 0.812 & 0.673 & 0.088 & 0.814 & 0.852 & 0.666 & 0.035 \\
LSR \cite{lv2021lsr_nc4k}      & CVPR$_{21}$ & 57.90  & 25.21  & 0.840 & 0.865 & 0.766 & 0.048 & 0.787 & 0.838 & 0.696 & 0.080 & 0.804 & 0.880 & 0.673 & 0.037 \\
TINet \cite{zhu2021tinet}      & AAAI$_{21}$ & 26.47  & 15.96  & 0.829 & 0.879 & 0.734 & 0.055 & 0.781 & 0.836 & 0.678 & 0.087 & 0.793 & 0.861 & 0.635 & 0.042 \\
C2FNet-V2 \cite{chen2022C2FNet-V2} & TCSVT$_{22}$ & 44.94 & 18.10 & 0.840 & 0.896 & 0.770 & 0.048 & 0.799 & 0.859 & 0.730 & 0.077 & 0.811 & 0.887 & 0.691 & 0.036 \\
PreyNet \cite{zhang2022preynet}    & MM$_{22}$    & 38.53 & 58.10 & 0.834 & 0.899 & 0.763 & 0.050 & 0.790 & 0.842 & 0.708 & 0.077 & 0.813 & 0.881 & 0.697 & 0.034 \\

SINet-V2 \cite{fan2021concealed}& TPAMI$_{22}$ & 26.98  & 12.28  & 0.847 & 0.903 & 0.770 & 0.048 & 0.820 & 0.882 & 0.743 & 0.070 & 0.815 & 0.887 & 0.680 & 0.037 \\
diffCOD \cite{chen2023diffCOD} & ECAI$_{23}$ & ---     & ---     & 0.837 & 0.891 & 0.761 & 0.051 & 0.795 & 0.852 & 0.704 & 0.082 & 0.812 & 0.892 & 0.684 & 0.036 \\
PFNet+ \cite{mei2023PFNet_plus} & SCIS$_{23}$ & 75.22 & 51.41 & 0.831 & 0.889 & 0.754 & 0.052 & 0.791 & 0.850 & 0.713 & 0.080 & 0.806 & 0.880 & 0.677 & 0.037 \\
WS-SAM \cite{he2024ws-sam} & NIPS$_{24}$  & ---     & ---     & 0.829 & 0.886 & \textbf{0.802} & 0.052 & 0.759 & 0.818 & 0.742 & 0.092 & 0.803 & 0.878 & \textbf{0.719} & 0.038 \\
\cmidrule(lr){1-16}
\multicolumn{15}{c}{\textbf{Efficiency Considered Methods}}\\
\cmidrule(lr){1-16}
ERRNet \cite{errnet}           & PR$_{22}$    & 69.76 & 20.05 & 0.827 & 0.887 & 0.737 & 0.054 & 0.779 & 0.842 & 0.679 & 0.077 & 0.786 & 0.867 & 0.630 & 0.043 \\
DGNet \cite{DGNet2023}         & MIR$_{23} $  & 21.02  & 2.77 & 0.857 & 0.911 & 0.784 & \textbf{0.042} & \textbf{0.839} & 0.901 & 0.769 & \textbf{0.057} & 0.822 & 0.896 & 0.693 & 0.033\\
\textbf{CS\(^3\)Net (Ours)} & --- & \textbf{20.52}  & \textbf{1.88}   & \textbf{0.859} & \textbf{0.915} & 0.792 & \textbf{0.042} & \textbf{0.839} & \textbf{0.903} & \textbf{0.774} & \textbf{0.057} & \textbf{0.825} & \textbf{0.901} & 0.703 & \textbf{0.032} \\ \bottomrule[1.5pt]
\end{tabular}
}
\label{tab:benchmark_result}
\vspace{-1em}
\end{table*} 
\subsection{Spotlight Shifting Strategy}
\label{subsec:spotlight_shifting_strategy}
\textbf{Motivation.} To achieve effective co-supervision within a single-branch framework, a straightforward yet information-dense prior must be devised. Drawing inspiration from the observation that shadow projection over an object undergoes eye-catching changes as the light source shifts its position, which makes blurred edge details easier to recognize. This concept serves as the basis for simulating the shadow projection effect in our algorithm.\\
\textbf{Shadow Projection Effect Simulation.} The shadow projection over an object is influenced by the position of the light source. To replicate this effect, we devise an algorithm that takes a ground truth map and selected points to produce shadow maps ($\mathcal{M}_{S}$), as illustrated in the left panel of Fig.~\ref{fig:network_struct}. Typically, shadow projections over an object form near its boundary, so the edge map (\(\mathcal{M}_{E}\)) is first extracted by computing the gradient of the ground truth map (\(\mathcal{M}_{GT}\)). To generate the shadow map based on a specific spotlight point \( q \) with coordinates \( (q_x, q_y) \),  every point \( p_i \in \mathcal{M}_{E} \) with 
coordinates \( (p_{i,x}, p_{i,y}) \) undergoes circular dilation \( \delta_{d_i}(\cdot) \), the dilation radius \( d_i \) is determined by the Euclidean distance between \( p_i \) and \( q \). \( \delta_{d_i}(\cdot) \) overlays a circle centered at \( p_i \) with the radius \( d_i \), scaled to a range of 0 to 30 before dilation. This function outputs a binary dilation map where the dilated area around \( p_i \) is marked as 1, and elsewhere as 0:
\begin{equation}
    \begin{array}{c}
    d_i(p_i, q) = \sqrt{(p_{i,x} - q_x)^2 + (p_{i,y} - q_y)^2} \quad \forall p_i \in \mathcal{M}_{E},\\
    \Delta(p_i) = \delta_{d_i(p_i, q)}(p_i) \quad \forall p_i \in \mathcal{M}_{E}.
    \end{array}
\end{equation}
To ensure that the desired shadow map \( \mathcal{M}_{S} \) is within the ground truth map \( \mathcal{M}_{GT} \), the following function is applied:
\begin{equation}
    \mathcal{M}_{S} = \mathcal{M}_{GT} \odot \left ( \bigcup_{p_i \in \mathcal{M}_{E}} \Delta(p_i) \right ),
\end{equation}
where \( \odot \) denotes element-wise multiplication, and \( \cup \) represents the union of all dilation results.\\
\textbf{Shadow Projection Feature Extraction.} Unlike other co-supervised networks \cite{edge_aware_SCRN,errnet,mutualGL,edge_reconstruion,edge_aware_mirrornet} with dual-branch architectures, we only use 4 layers of convolution to extract the shadow projection feature \( \Psi_{p} \) and predict the shadow map \( \mathcal{P}_s \), minimizing computational demand and parameters:
\begin{equation}
\label{equation:SRM}
    \Psi_{p} = Bconv^{iter=3}_{k=3}(x_1), \quad \mathcal{P}_s = Bconv^{iter=1}_{k=1}(\Psi_{p}).
\end{equation}
Here, \( Bconv^{iter}_{k}(\cdot) \) involves a \( k \times k \) convolution, batch normalization, and ReLU, repeated \( iter \) times.
\subsection{Projection Aware Attention}
The PAA module leverages the shadow projection feature \( \Psi_{p} \) and the concatenated output of the preceding PAA with $\left\{x_i\right\}_{i=2}^{4}$ to generate refined features $\left\{f_{i}\right\}_{i=1}^{4}$. This process uses ``chaotic mixing" ($Mix(\cdot,\cdot)$), which randomly concatenates and shuffles input channels. Following \cite{convattention2022restormer}, the PAA captures global and local contexts using 3$\times$3 depth-wise ($DConv(\cdot)$) and point-wise ($PConv(\cdot)$) convolutions:
\begin{equation}
\begin{array}{l}
    f_{1}=\Psi_{p}, \\
    f_{i}^{c} =\mathcal{D} ^{2}(f_{i-1})\oplus x_i , \\
    \mathcal{K}_i=\mathcal{V}_i=DConv(PConv(f_{i}^{c})), \\
    \mathcal{Q}_i= DConv(PConv (Mix(\mathcal{D} ^{2^{i-1}}(\Psi_{p} ),f_{i}^{c}))), \\
    f_{i}=PConv\left (softmax\left ( \frac{\mathcal{Q}_i\mathcal{K}_i^T}{\alpha} \right )  \mathcal{V}_i \right ), \text{ for } i = 2,3,4, \\
\end{array}
\end{equation}
where $\mathcal{D}^{n}(\cdot)$ is $n$-fold down-sampling, $\oplus$ indicates channel concatenation, and $\alpha$ is a learnable scaling factor.
\subsection{Extended Neighbor Connection Decoder}
Building on the idea of neighbor connection \cite{fan2021concealed}, we extend an extra path for high-resolution feature, which is normally discarded yet crucial for detail reconstruction. The ENCD combines multi-scale features $\left \{ f_{i} \right \} _{i=1}^{4}$ while maintaining semantic consistency. These features are transited to $ N_C$ channels via point-wise convolution before ENCD. As shown in the bottom right of Fig.~\ref{fig:network_struct}, the feature aggregation strategy can be expressed as:
\begin{equation}
\left\{     
\begin{array}{l}
    f_{4}^{'}=f_{4},  \\
    f_{3}^{'}=f_{3}\odot \mu^2(Bconv_{3}(f_{4})), \\
    f_{2}^{'}=f_{2}\odot \mu^2(Bconv_{3}(f_{3}))\odot\mu^2(Bconv_{3}(f_{3}^{'})),\\
    f_{1}^{'}=f_{1}\odot \mu^2(Bconv_{3}(f_{2}^{'})), \\
\end{array}
\right .
\end{equation}
where $f_{i}$ and $f_{i}^{'}$ denote the features before and after the neighbor connection. $\mu^2(\cdot)$ and $\odot$ represent two times up-sampling and element-wise multiplication respectively. Using the refined features $\left \{ f{'}_{i}  \right \} _{i=1}^{4}$, the decoding process is:
\begin{equation}
\begin{array}{l}   
    \mathcal{P}_{GT} = Bconv_{1}(\mathcal{F}(\mathcal{F}(\mathcal{F}(f_{4}^{'}, f_{3}^{'}), f_{2}^{'}), f_{1}^{'})), \\
    \text{where }\mathcal{F}(x, y) =Bconv_{3}(Bconv_{3}(\mu^2(x)) \oplus y).
\end{array}
\end{equation}
In this context, \(\mathcal{P}_{GT}\) represents the final prediction, and \(\oplus\) denotes channel-wise concatenation.
\subsection{Loss Function and Co-supervision}
\label{subsec:co-sup}
As the algorithm described in Sec.~\ref{subsec:spotlight_shifting_strategy}, different shadow maps are synthesized into a unified one via pixel-wise addition. This map serves as a critical supervisory element, informing the calculation of the shadow map loss \( \mathcal{L}_\text{S} \) by assessing the mean squared error (MSE) between \( \mathcal{P}_S \) and \( \mathcal{M}_{S} \).
\par The total loss \( \mathcal{L} \) is a combination of \( \mathcal{L}_\text{S} \) and the ground truth loss \( \mathcal{L}_{GT} \). The latter consists of the weighted Intersection over Union (IoU), \( L^{w}_\text{IoU} \), and weighted binary cross-entropy (BCE), \( L^{w}_\text{BCE} \), as used in the literature \cite{COD10K_SINet,fan2021concealed,DGNet2023,wei2020f3net}. The complete loss function can be expressed as:
\begin{equation}
    \mathcal{L} = \mathcal{L}_{GT}(\mathcal{P}_{GT}, \mathcal{M}_{GT}) + \mathcal{L}_\text{S}(\mathcal{P}_{S}, \mathcal{M}_{S}).
\end{equation}

\begin{figure*}[th]
    \centering
    \includegraphics[width=\textwidth]{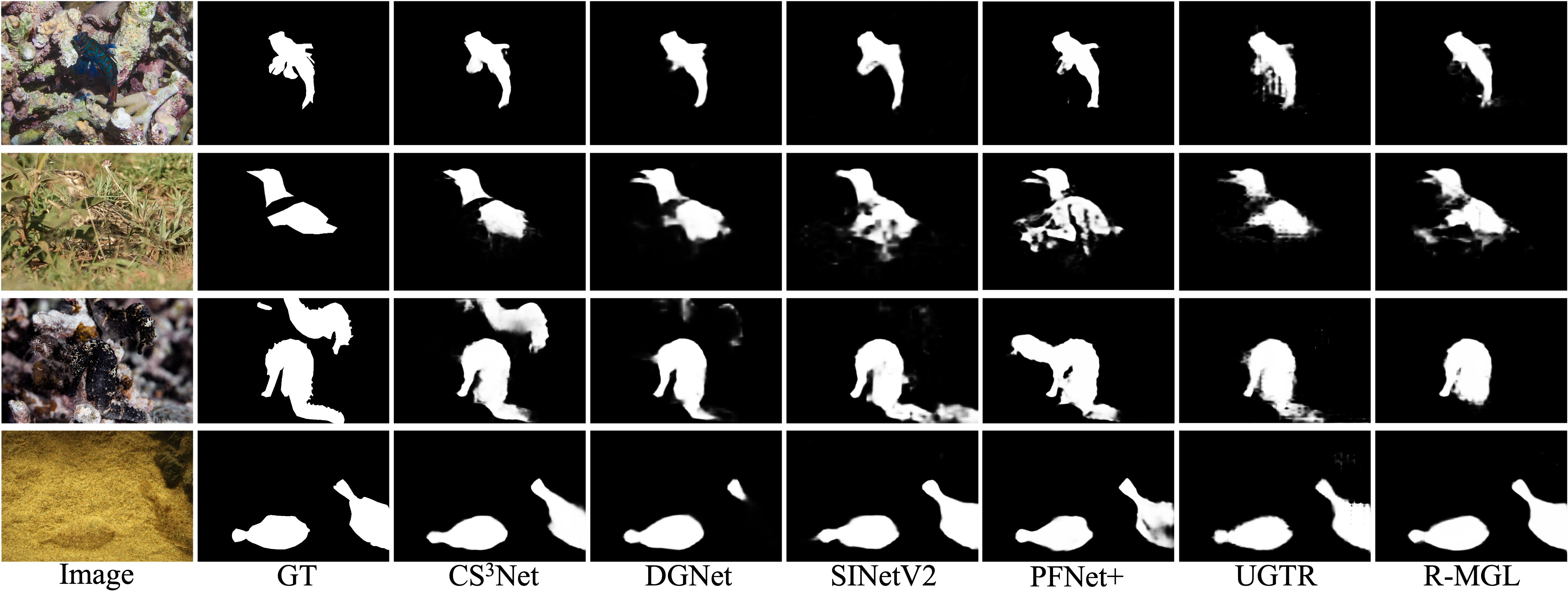}  
    \vspace{-2.7em}
    \caption{Visual performance of CS\(^3\)Net and other highly competitive COD methods.}
    \label{fig:qualiatative_result}
    \vspace{-1.3em}
\end{figure*}

\section{Experiments And Results}
\label{sec:experiments}

\subsection{Experiment Settings} \label{subsec:experiment_settings}
\textbf{Datasets.} In line with most COD methods, our model is trained on the COD10K \cite{COD10K_SINet} and CAMO \cite{CAMO_Abranch} datasets, and tested on the NC4K \cite{lv2021lsr_nc4k}, COD10K, and CAMO datasets. CAMO contains 1,250 images, with 1,000 used for training and 250 for testing. COD10K consists of 5,066 images, with 3,040 for training and 2,026 for testing. NC4K is a challenging test dataset comprising 4,121 images.
\\
\textbf{Metrics.} Following previous works \cite{fan2021concealed,yang2021UGTR,lv2021lsr_nc4k,mutualGL}, four widely used metrics are employed for evaluation:
Mean Absolute Error (MAE), Mean E-measure ($E_\phi$) \cite{fan2021emeasure}, S-measure ($S_\alpha$) \cite{fan2017smeasure}, and Weighted F-measure ($F^w_\beta$) \cite{margolin2014fmeasure}. To assess the efficiency of our method\footnote{The model efficiency is ascertained using the flops-counter toolbox: {https://github.com/sovrasov/flops-counter.pytorch}}, the number of model parameters (in Millions, M) and the quantity of multiply-accumulate (MACs) operations (in Giga, G) are used as indicators. \\
\textbf{Implementation Details.} Our model is trained on a single NVIDIA RTX-3090 GPU. The training settings including optimizer, image resizing and data augmentation mirror those found in \cite{DGNet2023}. The PAA is configured with 4 heads ($N_H$) and the $N_C$ of ENCD is set to 32. The shadow map is generated using the spotlight shifting strategy, with the top-left and bottom-right of the image preset as spotlight points. \\
\begin{figure}[htbp]
    \centering
    \vspace{-1.7em}
    \includegraphics[width=\linewidth]{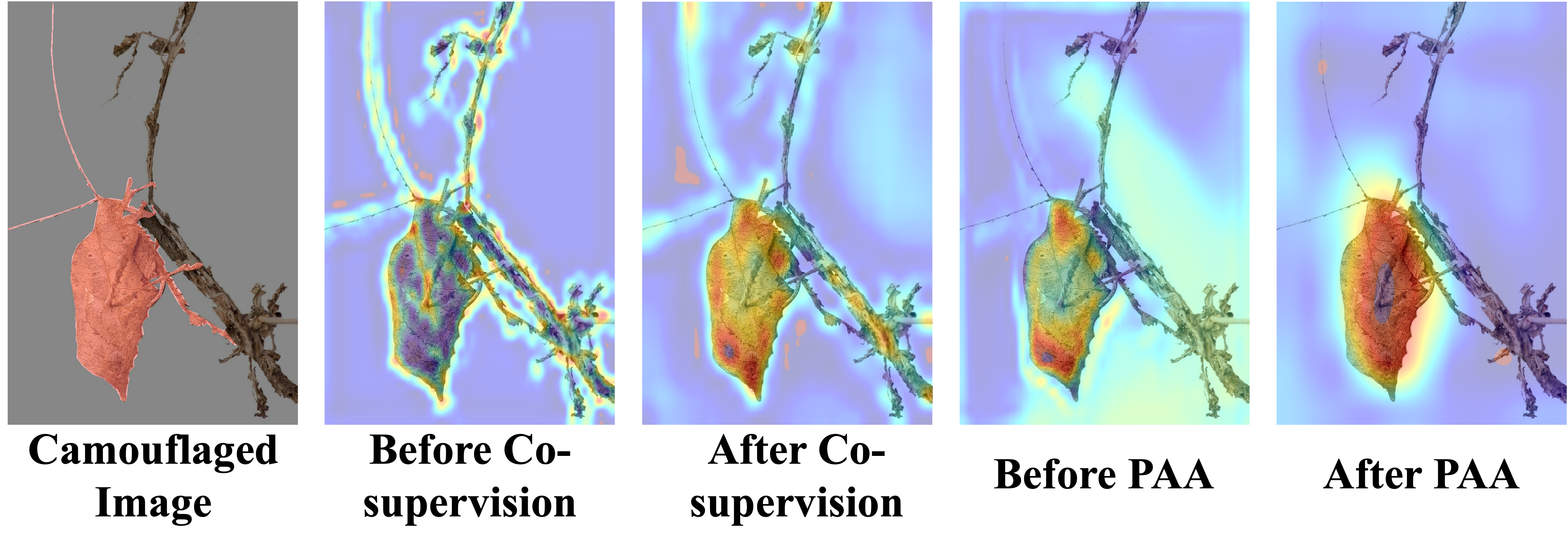} 
    \vspace{-2.6em}
    \caption{Visualization of feature maps before and after proposed modules: Spotlight Shifting Co-supervision and PAA.}
    \label{fig:feature_viz}
    \vspace{-1.3em}
\end{figure} 
\subsection{Comparison with the State-of-the-Art Methods} \label{subsec:comparison}
Our CS\(^3\)Net is benchmarked against recent SOTA methods, including 2 efficiency focused methods. For fair competition, all results are derived either directly from the associated publications or by reproducing with the default configurations. As presented in Tab.~\ref{tab:benchmark_result}, our model demonstrates exceptional performance across all datasets with a notable reduction in MACs (32.13\% drop compared to the leading efficiency focused method, DGNet \cite{DGNet2023}) and smaller model size. Specifically, CS\(^3\)Net achieves higher $S_\alpha$ score than SINet-V2 \cite{fan2021concealed} by 2.86\% in NC4K and improves $F^w_\beta$ score by 3.38\% in COD10K. Fig.~\ref{fig:qualiatative_result} clearly demonstrates the superiority of our model in accurately segmenting multiple camouflaged objects and those partially occluded by surrounding elements.\\
\begin{table}[th]
    \vspace{-0.9em}
    \caption{Ablation study on the CAMO dataset.}
    \vspace{-0.9em}
        \centering
        \resizebox{\linewidth}{!}{%
        \begin{tabular}{llcccc}
        \toprule[1.5pt]
        \multicolumn{2}{c}{\multirow{2}{*}{Method}}               & \multicolumn{4}{c}{CAMO}              \\ 
        \cmidrule(lr){3-6}
        \multicolumn{2}{c}{}                                      & \textbf{$S_\alpha$ ↑}& \textbf{$E_{\phi}$ ↑}& \textbf{$F^w_\beta$ ↑}& \textbf{$MAE$ ↓}\\
        \cmidrule(lr){1-6}
        \multicolumn{2}{l}{Baseline}                              & 0.798    & 0.804  & 0.643     & 0.084 \\
        \multicolumn{2}{l}{Baseline+ENCD}                         & 0.829    & 0.889  & 0.761     & 0.061 \\
        \multicolumn{2}{l}{Baseline+ENCD+PAA}                  & \multirow{2}{*}{0.832}    & \multirow{2}{*}{0.890}  & \multirow{2}{*}{0.752}     & \multirow{2}{*}{0.062} \\
        \multicolumn{2}{l}{(w/o Co-supervision)}                  &     &   &      &  \\
        \multirow{3}{*}{Our Method} & Edge Map     & 0.835    & 0.898  & 0.768     & 0.059 \\
                                    & Dilated Edge & 0.835    & 0.897  & 0.760     & 0.061 \\
                                    & Spotlight Shifting & \textbf{0.839}    & \textbf{0.903}  & \textbf{0.774}     & \textbf{0.057}
        \\ \bottomrule[1.5pt]
        \end{tabular}%
        }
        \label{tab:ablation_study}
        \vspace{-1.8em}
\end{table}
\vspace{-1.6em}
\subsection{Ablation Study} \label{subsec:ablation_study} 
Tab.~\ref{tab:ablation_study} presents our ablation study, where we also compare the effects of different co-supervision strategies on our method. Here, the term ``baseline" refers to the EfficientNet-B4 with convolution layers to replace other modules.\\
\textbf{Effectiveness of ENCD.} Tab.~\ref{tab:ablation_study} demonstrates that the ENCD module is crucial for enhancing the model's performance, as evidenced by the significant improvement in all metrics.\\
\textbf{Effectiveness of PAA.} PAA alone shows limited improvement as it relies on the integration of informative cues from co-supervision. However, when combined with co-supervision, it significantly boosts performance (last 3 rows of Tab.~\ref{tab:ablation_study}), Fig.~\ref{fig:feature_viz} further shows that the feature map highlights the camouflaged objects more distinctly when PAA is applied. \\
\textbf{Effectiveness of Spotlight Shifting.} Tab.~\ref{tab:ablation_study} demonstrates that the spotlight shifting strategy provides superior performance compared to using edge or dilated edge co-supervision. Specifically, spotlight shifting achieves the best results across all key metrics. And it effectively suppresses irrelevant features and highlights the essential regions, as shown in Fig.~\ref{fig:feature_viz}.\\
\vspace{-2em}
\section{Conclusion} \label{sec:conclusion}
Our research introduces a novel single-branch efficient network, CS\(^3\)Net for camouflaged object detection. The model leverages the spotlight shifting strategy to simulate the shadow projection effect, enhancing the visibility of camouflaged objects. And we proposes PAA for feature aggregation and ENCD for efficient decoding. Extensive experiments show that CS\(^3\)Net strikes the optimal balance between performance and efficiency, achieving a 32.13\% reduction in MACs compared to the cutting-edge efficient COD methods, while enhancing detection accuracy, indicating a step
forward in performance-conscious COD model design.

\bibliographystyle{IEEEtran}
\bibliography{references}{}
\end{document}